# BEEM : Bucket Elimination with External Memory

**Kalev Kask, Rina Dechter and Andrew E. Gelfand**
Donald Bren School of Information and Computer Science
University of California, Irvine

## Abstract

A major limitation of exact inference algorithms for probabilistic graphical models is their extensive memory usage, which often puts real-world problems out of their reach. In this paper we show how we can extend inference algorithms, particularly Bucket Elimination, a special case of cluster (join) tree decomposition, to utilize disk memory. We provide the underlying ideas and show promising empirical results of exactly solving large problems not solvable before.

## 1 Introduction

Exact inference algorithms for graphical models broadly fall into two categories: 1) Inference-based algorithms (e.g. Bucket Elimination (BE), Cluster Tree Elimination (CTE)); and 2) Search-based (e.g. best-first, depth-first branch and bound) [3, 4, 5, 2]. Inference-based algorithms are time exponential in the induced width and also require space exponential in the induced-width. While brute force search algorithms can work in linear space (e.g., depth first search), more advanced search schemes that use and/or search spaces require memory exponential in the induced-width as well. Consequently, both classes of algorithms are feasible only when the induced-width (or treewidth) is small due to memory limitations.

Not surprisingly, various algorithms have been proposed that can work with bounded memory at the expense of additional time [2, 8, 9, 10]. In this paper we aim to push the boundary of memory-intensive algorithms further, allowing a more effective tradeoff or in some cases eliminating the compromise altogether. To do so, we extend the memory available to the algorithm to include external (disk) memory.

In comparison to main/virtual memory, external memory is seemingly unlimited. So an algorithm that effectively utilizes external memory should, in principle, be able to tackle problems with very large induced widths. However, the additional space does not come without cost, as access of disk memory is typically orders of magnitude slower than main memory. Nonetheless, it has been demonstrated in the context of (A*) heuristic search that algorithms can be designed to mitigate such effects [6, 7], yielding powerful schemes that can be applied to previously unsolvable problems.

To make the realm of problems solvable by inference algorithms using external memory more concrete, consider a Bayesian network (BN) comprised of binary variables ($k = 2$) and having induced width, $w^\star = 20$. The largest table in this model has $2^{19} = 524,288$ entries. Assuming that a double-precision floating point number requires 8 bytes, this problem requires about 4MB of memory and easily fits into main memory. However, if each variable in the model is ternary ($k = 3$) rather than binary, the largest table requires about 9GB of memory. While 9 GB is more main memory than most computers have, such a problem would fit comfortably into external disk, which can exceed several Tera-bytes in size.

In the remainder of this paper, we describe how a specific inference-based algorithm, BE [3], can be modified to use external memory. We demonstrate the performance of this new algorithm, named Bucket Elimination with External Memory (BEEM), for computing the probability of evidence on a class of large networks for which exact computations had not previously been made, and otherwise show that it is faster then some of the best algorithms that trade space for time.

## 2 Background

In this section we present some necessary preliminaries on graphical models and Bucket Elimination.

DEFINITION **1.** *Graphical Model* - *A Graphical model $\mathcal{R}$ is a 4-tuple $\mathcal{R} = \langle X, D, F, \otimes \rangle$, where:*

1. $X = \{X_1, ..., X_n\}$ is a set of variables;
2. $D = \{D_1, ..., D_n\}$ is the set of their respective finite domains of values;
3. $F = \{f_1, ..., f_r\}$ is a set of real valued functions defined over a subset of variables $S_i \subseteq X$. The scope of function $f_i$, denoted $scope(f_i)$, is its set of arguments, $S_i$;
4. $\otimes_i f_i \in \{\prod_i f_i, \sum_i f_i, \bowtie_i f_i\}$ is a combination operator.

The graphical model represents the combination of all its functions: $\otimes_{i=1}^{r} f_i$.

The primal graph of a graphical model associates a node with each variable and connect any two nodes whose variables appear in the same scope.

DEFINITION 2. *Induced Width* - An ordered graph is a pair $(G, d)$, where $G$ is an undirected graph and $d = X_{(1)}, ..., X_{(n)}$ is an ordering of the nodes ($X_{(i)}$ means the $i^{th}$ node in the ordering). The width of a node is the number of the node's neighbors that precede it in the ordering. The width of an ordering $d$ is the maximum width over all nodes. The induced width of an ordered graph $w^\star(d)$ is the width obtained when nodes are processed from last to first, such that when node $X_{(i)}$ is processed, all of its preceding neighbors ($X_{(j)}$ for $j < i$) are connected. The induced width of a graph, $w^\star$, is the minimal induced width over all possible orderings.

**Bucket Elimination (BE)** is a special case of cluster tree elimination in which the tree-structure upon which messages are passed is determined by the variable elimination order used [3]. In BE terminology, the nodes of the tree-structure are referred to as buckets and each bucket is associated with a variable to be eliminated. Each bucket contains a set of functions, either the original functions (e.g. Conditional Probability Tables (CPTs) in a BN), or functions generated by the algorithm. Each bucket is processed by BE in two steps. First, all functions in the bucket are combined (by multiplication in the case of BNs ). Then the variable associated with the bucket is eliminated from the combined function (by summation in case of Belief Updating in BNs). The function resulting from the combination and elimination steps is then passed to the parent of the current bucket. Processing occurs in this fashion, from the leaves of the tree to the root, one node (bucket) at a time as illustrated in Figure 1. *It is important to note that the bucket-tree induces a partial order on the variables in which child nodes (variables) are processed prior to their parents.*

Formal definitions of BE data-structures are given next for completeness. A formal description of the BE algorithm is also presented in Figure 2.

DEFINITION 3. **Bucket** Let $B_{x_1}, ..., B_{x_n}$ be a set of buckets, one for each variable and let $d$ be ordering of these variables. Each bucket $B_{x_i}$ contains those functions in $F$ whose latest variable in $d$ is $X_i$.

DEFINITION 4. **Bucket Tree** Let $G_d^\star$ be the induced graph along an ordering $d$ of a reasoning problem whose primal graph is $G$. The vertices of the bucket-tree are the $n$ buckets, which are denoted by their respective variables. Each vertex $B_X$ points to $B_Y$ (or, $B_Y$ is the parent of $B_X$) if $Y$ is the latest neighbor of $X$ that appear before $X$ in $G_d^\star$. The degree of bucket $B$, denoted $deg_B$, is the number of neighbors of bucket $B$ in the bucket-tree.

DEFINITION 5. **Input-Output Functions** Given a directed bucket-tree $T$, for any bucket $B$, the output function of $B$ is the function that $B$ sends to its parent, and the input functions of $B$ are functions that $B$ receives from its children.

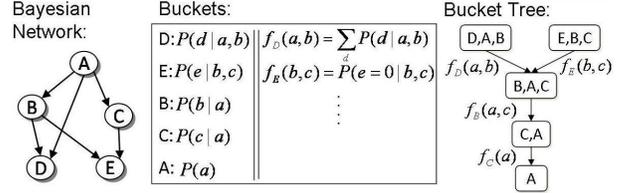

Figure 1: Illustration of BE for the query $P(a|e = 0)$ given ordering $d = \{A, C, B, E, D\}$. Bucket D is processed first. The CPT $P(d|a, b)$ is placed in bucket D and a function $f_D(a, b)$ is generated and passed to bucket B. Next bucket E is processed $f_E(b, c)$ is sent to bucket B. Then bucket B is processed by multiplying $P(b|a)$, $f_D(a, b)$ and $f_E(b, c)$ and eliminating variable B. Processing on the bucket tree continues in this manner until the query can be computed.

## 3 Bucket Elimination with External Memory

Processing a bucket in a bucket-tree involves two operations: 1) Combining functions in the bucket; and 2) Eliminating the bucket's variable[1]. In many problems, the function resulting from these operations is too large to fit into memory. For example, assume we are computing a function $f_{X_p}$, by eliminating a variable $X_p$ from a function $h$. Normal BE (as defined in Figure 2) cannot operate if the function $h$ does not fit into main memory in its entirety. Since computers have orders of magnitude more disk space than main memory, a straightforward modification of BE would be to divide large tables (such as $h$) into blocks that fit into main memory and store these blocks to disk.

---
[1]In the remainder of this paper, we assume that functions take a tabular form.

> Algorithm **Bucket Elimination for P(e)**
> **Input:** A problem description $P = <X, D, F>$; evidence variables $var(e)$; and an ordering of the variables $d = (X_{(1)}, ..., X_{(n)})$.
> **Output:** Probability of Evidence p(e).
>
> • **Initialize:** Partition the functions in F into buckets denoted $B_{X_1}, ..., B_{X_n}$, where initially $B_{X_i}$ contains all input functions whose highest variable is $X_i$ (ignore instantiated variables).
>
> During the algorithm's execution $B_{X_i} = f_1, f_2, ..., f_j$
> • **Backward:** For $p \leftarrow n\ downto\ 1$, process $B_{X_i}$:
>
> - Generate the function $f_{X_p}$ by:
>   1) Combination: $f'_p = \prod_{f \in B_{X_p}} f$
>   2) Elimination: $f_{X_p} = \sum_{X_p} f'_p$
>
> - Add $f_{X_p}$ to the bucket of the largest-index variable in the scope of $f_{X_p}$.
>
> • **Return:** $p(e) = \prod_{f \in B_{X_1}} f$

Figure 2: The Bucket Elimination Algorithm [3]

The function $f_{X_p}$ can then be computed by loading required blocks from hard disk. In an extreme case, one could compute $f_{X_p}$ one entry at a time, each time loading the relevant entries of the input function $h$ from disk, while saving entries of $f_{X_p}$ as they are computed.

The performance of this naive, entry by entry, algorithm would be extremely poor because:

- While main memory has bandwidth (data transfer rate) of a few GB/second, disk memory typically has sequential transfer rate of 100 MB/second, and much worse non-sequential transfer rate.
- Main memory has 0 seek time (since it allows random access), while disk memory has a seek time of about 10 ms.

As a result, this naive algorithm would spend most of its time waiting for table entries to be loaded and then saving them (i.e. in disk I/O), rather than performing actual computations. The time spent on disk I/O has a linear component that depends on block size and a fixed component given by the seek time. In most applications the fixed component dominates, suggesting that the primary goal in designing a disk-based adaptation of any algorithm is to minimize the number of reads/writes from hard disk.

The challenge of minimizing disk reads/writes is further compounded in a multi-threaded environment. Since individual entries of a function table are independent, function blocks can be processed in parallel. Thus, while one thread is waiting for data to be loaded from disk, other threads can carry out computation on their assigned blocks. This parallelism offers the potential for algorithmic speed-up, while at the same time introducing a new scheduling challenge. Specifically, one now has to schedule disk I/O so as to minimize the amount of time that each thread waits for data to be loaded/saved.

In this paper, we address the two, potentially conflicting goals of minimizing reads/writes and limiting thread starvation, by decomposing the challenge into two tasks: 1) Function table indexing; and 2) Block-size computation. The block size computation task involves dividing the function tables into blocks that are as large as possible. The function table indexing task involves arranging the entries within a table (and block) so as to minimize the number of reads/writes to disk. Both of these tasks must be addressed within the constraints imposed by the bucket-tree. The following two subsections describe our approach to these two design aspects.

### 3.1 Function Table Indexing

When processing a table, all of the table's entries are ordered, assigned an index consistent with that order, and then processed one-by-one. For example, if $f(X_1, X_2)$ is a table of ternary variables, the entries are ordered as: $<0, 0>$ (index 0), $<0, 1>$ (index 1), $<0, 2>$ (index 2), $<1, 0>$ (index 3), etc. The ordering of variables in the scope of a function thus dictates where an entry is located within that function's table. Since we are considering functions (i.e. $f_{X_p}$) that are broken into blocks because they do not fit into memory, the ordering of variables can also impact the number of reads/writes to disk.

In the following section, we illustrate how the ordering of variables within a scope can impact the performance of our algorithm. We then show how some of these inefficiencies are remedied by the scope ordering imposed by the bucket-tree structure. Since processing a bucket and generating its output function involves two steps - combination and elimination - we analyze the ordering constraints imposed by these two steps.

#### 3.1.1 Ordering Constraints due to Elimination

Assume we are computing a function $f(X_1, X_2)$, by eliminating variable $Y$ from the function $h(Y, X_2, X_1)$ (as shown in Figure 3). Furthermore, assume that we are computing entry 1 (corresponding to value combination $<0, 1>$) of $f$. Since $f(X_1 = 0, X_2 = 1) = \sum_Y h(Y, X_2 = 1, X_1 = 0)$, we need entries 3, 12, 21 from table $h$ (corresponding to argument combinations $<0, 1, 0>$, $<1, 1, 0>$, $<2, 1, 0>$, respectively). As mentioned in the previous section, data is loaded in predetermined size blocks, each stored as a separate file. If the blocks of $h$ were only 8 entries in size, the required entries (3, 12, 21) would reside in different blocks and three block load operations would be required. When processing the next entry of $f$ (corresponding to value combination $<0,2>$), we then require entries 6, 15, 24, which also reside in different blocks. Worse yet, since only one block per table is kept in memory at a time, the blocks of $h$ containing entries 6, 15, 24 were in memory when entry 1 of $f$ was computed, but then unloaded to make room for subsequent blocks. Thus, many unnecessary block loads/unloads are performed due to the poor ordering of variables in the scope. Our goal is to minimize the number of times any block is loaded.

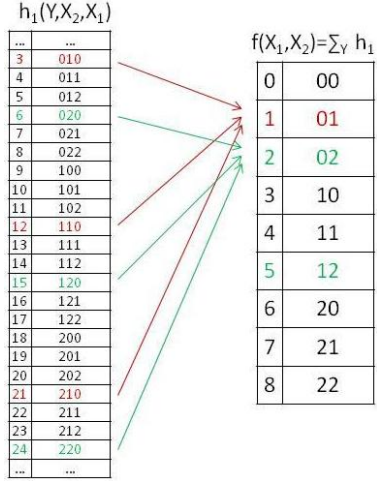

Figure 3: Function Table Indexing Example

More generally, computing a function $f$ by eliminating a variable from function $h$ involves maintaining two lists of table entries. First, we enumerate the entries of $f$ and then we identify the entries of $h$ required by each entry of $f$. A condition sufficient to guarantee that every block is loaded no more than once is that the enumeration of entries of $h$ be monotonically organized. This condition was not satisfied in the previous example, since to compute entries 1 and 2 of $f$ we needed entries 3, 12, 21 and 6, 15, 24 of $h$, respectively.

DEFINITION 6 (monotone computation of $h$ relative to $f$). If $f$ is obtained from $h$ by eliminating $X$ ($f = \sum_x h$), we say that $h$ is monotone relative to $f$ iff: 1. the computation of a single entry of $f$ requires only consecutive entries of $h$, 2. the computation of successive entries in $f$ requires successive collections of consecutive entries in $h$.

Clearly the non-monotoncity stems from the fact that the order of variables in the scopes of $f$ and $h$ do not agree. By rearranging the scope of $h$ to $<X_1, X_2, Y>$, we can make the enumeration of $h$ monotonic wrt $f$. In particular, the entries of $h$ needed to compute $f(X_1 = 0, X_2 = 1)$ are at indices 3, 4, 5, and the entries needed to compute $f(X_1 = 0, X_2 = 2)$ are at indices 6, 7, 8. Under this ordering, no redundant loading/unloading of blocks of $h$ are required. The monotonicity of $h$ relative to $f$ can thus be achieved by the following two ordering constraints:

**Proposition 1.** *Let $f$ be obtained from $h$ by eliminating $X$,*

1. *If $X$ is the last variable in the scope of $h$, and*
2. *If the order of the remaining variables in the scope of $h$ agrees with the order of the variables in the scope of $f$,*

*then $h$ is monotone relative to $f$.*

*Proof.* The first condition guarantees that the entries of the input function $h$ needed for the computation of any entry of the output function $f$ are consecutive in the table of $h$. The second condition ensures the second requirement of monotonicity. □

It turns out that the constraints for monotonicity can be satisfied over all buckets, simultaneously. In fact, the topological, partial ordering dictated by the bucket tree (from leaves to root) can be shown to guarantee monotonicity.

**Proposition 2.** *Given a directed bucket-tree $T = (V, E)$, where $V$ are the bucket variables and $E$ are the directed edges, any partial order along the tree and, in particular the order of bucket-elimination execution, yields monotonic processing of buckets relative to elimination.*

Since the variables in the scope of each function correspond to nodes in the bucket-tree that have yet to be processed, we can infer the following.

**Proposition 3.** *The partial order along the bucket-tree yields a strict ordering within each function scope.*

Algorithm **Scope Ordering** takes as input a bucket-tree and orders the scopes of functions in all buckets, according to the topological ordering of the given bucket-tree. It works by traversing the bucket tree from the root in a breadth-first manner and at each bucket ordering the scopes of the input functions wrt the bucket's output function.

### 3.1.2 Ordering Constraints due to Combination

Our discussion thus far has focused on the elimination step. During the combination step, a new set of ordering constraints are needed to ensure monotonicity. To illustrate, consider the following example. Assume we are computing a function $f(X_1, X_2, X_3)$, by combining and eliminating variable $Y$ from functions

$h_1(X_1, X_2, Y)$, $h_2(X_1, X_3, Y)$, $h_3(X_2, X_3, Y)$. Further assume that the domain size of all variables is 3. From the previous section, we know that $Y$ must be the last variable in the scope of the combined function. Thus, when computing $f(X_1 = 1, X_2 = 1, X_3 = 2)$, we need to access entries $h_2(X_1 = 1, X_3 = 2, Y = 0, 1, 2)$, corresponding to indices 15, 16 and 17 in table $h_2$. The next entry of $f(X_1 = 1, X_2 = 2, X_3 = 0)$, requires entries $h_2(X_1 = 1, X_3 = 0, Y = 0, 1, 2)$, corresponding to indices 9, 10 and 11. Finally, the entry $f(X_1 = 1, X_2 = 2, X_3 = 1)$ requires entries $h_2(X_1 = 1, X_3 = 1, Y = 0, 1, 2)$, corresponding to indices 12, 13 and 14. This enumeration of $h_2$ is non-montonic, even though it is consistent with the constraints imposed by elimination.

The problem occurs because the scope of $f$ has a 'gap' with respect to the scope of $h_2$. Specifically, $f$ has a variable $X_2$ between variables $X_1$ and $X_3$ that is not in the scope of $h_2$. In this example, there is no ordering of the variables that will avoid such a 'gap'. In many situations, such gaps occur due to ordering constraints imposed by the elimination order.[2]

### 3.2 Block Computation

As discussed earlier, we assume that all tables of intermediate functions are handled (i.e. loaded, computed, saved) as blocks. Clearly, loading/saving one table entry at a time is inefficient. Intuitively, then, we should divide function tables into blocks that are as large as possible. However, block size is limited by several factors. First, our algorithm is operating in a shared memory setup and each thread requires memory to operate. In addition, processing a bucket requires enough space in memory for the output function block and the blocks of each input function. Our goal is thus to determine how to divide function tables into blocks that minimize unutilized memory.

To simplify this problem we make the following design assumptions:

1. We assume the original functions occupy little space and can be stored in memory at all times;[3]

2. We assume that each thread uses the same amount of memory and that the memory allocated to a thread remains fixed throughout the algorithm's execution;

3. We assume that each bucket's output function table is broken into equally sized blocks; and

4. When computing a block of an output function, a thread requires enough space in memory for the output function block and the necessary blocks from each input function. That is, to compute a block of $f_{X_p} = \sum_{X_p} \prod_{f \in B_{X_p}} f$, we assume that the needed blocks of $f \in B_{X_p}$ are in memory.

The first two assumptions imply that the amount of memory available to each thread, denoted by $Mpt$, is $Mpt = (M - O)/m$, where $M$ is the total memory available, $O$ is the memory occupied by the original functions and $m$ is the number of threads. It is worth noting that the third design restriction does not imply that all block sizes are the same size.

Under the above design restrictions, a workable and simple upper bound on block size of all the functions residing in a bucket $B_X$ is $Mpt/deg_{B_X}$ - i.e. allocate the memory equally among the output function of a block and the output functions of its children in the bucket tree. Since each block is used twice - once as an output block and once as input block - and the degree of the buckets operating on a block may (and most likely will) be different, we need to coordinate this upper-bound between adjacent buckets.

To illustrate this issue, consider the function $f_u$ sent by bucket $u$ to bucket $v$ in Figure 4. When function $f_u$ is computed, its bucket imposes an upper bound on its block size as $Mpt/3$, since its degree is 3. However, when $f_u$ is used by parent bucket, $v$, its block size is bounded by $Mpt/6$, since bucket $v$'s degree is 6. Therefore, setting the block size of $f_u$ to $Mpt/3$ equally among all the bucket's function block would violate the fourth design restriction when bucket $v$ is processed.

The limitations on block sizes can be captured more formally by the following set of simultaneous constraints. Given a bucket tree and the root node of the bucket tree, the block size of each bucket must satisfy

$$Mpt = \epsilon_i + b_i + \sum_{j \in C(X_i)} b_j \quad \text{for each bucket } i \quad (1)$$

where $b_i$ is the block size for bucket $i$, $C(X_i)$ is the set of children of bucket $i$ and $\epsilon_i$ is the unutilized space in the computation of blocks for bucket $i$.

With the block size constraints in place, the problem of minimizing the amount of unutilized memory can be formalized as

$$\{b_1^\star, ..., b_n^\star\} = \arg\min_{\{b_1,...,b_n\}} \sum_{i=1}^{n} \epsilon_i \quad (2)$$

$$\text{s.t. Eqn.1}, \quad b_i \geq 0, \quad \epsilon_i \geq 0, \forall i$$

---

[2]The removal of 'gaps' is untreated at this point and we are currently exploring a variety of approaches for dealing with this issue.

[3]This assumption is for simplicity. The original function could also be broken into blocks and stored on the disk.

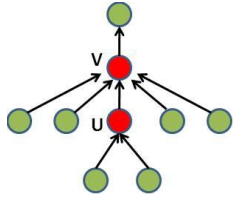

Figure 4: Block Size Computation for Function $f_u$

where $Mpt$ in Eqn.1 is a constant and the children of each node are governed by the underlying bucket-tree structure.

Eqn. 2 is a standard constraint satisfaction problem. In Figure 5 we provide a greedy algorithm that provides a feasible, though possibly suboptimal, solution to this problem. The algorithm computes block sizes starting with the buckets of highest degree (since they are the most constrained) and continues processing buckets in decreasing order of degree. At each bucket the remaining memory (i.e. the memory not already allocated to functions in that bucket) is divided equally between the undetermined functions in that bucket.

**Proposition 4.** *The complexity of the Block Size Computation algorithm is $O(n \cdot log(n))$, where $n$ is the number of variables.*

It is worth noting that many different block sizes can satisfy the constraints in Eqn.1 depending on which variable is chosen as the root of the tree. This is why we require the bucket-tree structure to be fixed in advance. It is also worth noting that $b_i$, the block size for bucket $i$, is not a continuous variable; rather it is some multiple of the operating system's cluster size. However, in practice we have found this relaxation to be non-problematic.

### 3.3 The BEEM Algorithm

We have developed a new algorithm, called BEEM, that incorporates the design ideas and algorithms presented in this section. The basic outline of the BEEM algorithm is given in Figure 6.

There is a 1-to-1 mapping between blocks and files. A file name is a concatenation of the bucket's variable (that generated the function that the block belongs to) and the block index (wrt the function table). For example, if the function generated by bucket $B_X$ is split into 5 blocks, the 5 files that contain the data will be named "X-1", "X-2", "X-3", "X-4", "X-5". When a particular block is needed, the program looks for and loads/saves a file with the appropriate name. A block is an array of double-precision floating-point numbers, and entire block/file can be loaded/saved with a single command.

---

Algorithm **Block Size Computation**

**Input:** A bucket tree, with buckets $B_{X_i}, ... B_{X_n}$ along ordering $d$.
**Output:** A set of block sizes, denoted $b_i$, for each output function $f_{X_i}$ of bucket $B_{X_i}$.

- **Do** i = 1 to n, in decreasing order of degree in the bucket-tree:

  1. Let $B_X \leftarrow B_i$; let $I$ be the indices of child variables of $X$ in the bucket-tree whose block size was not yet determined and $J$ those indices whose block size was already determined.

  2. For each undetermined function $f_{X_j}$ for $j \in I$ compute its block size as
  $$b_j \leftarrow \frac{Mpt - \sum_{k \in J} b_k}{deg_{B_X} - |I|}$$

Figure 5: Algorithm for computing block size

A basic step of the algorithm is computing a function table block. First we pick the block to compute, from an eligible bucket (*Computation.1*). A block is computed by enumerating all entries as described in *Computation.2*. In particular, we determine the indices of the entries in each input table and output table as described in the Function Table Indexing section. Based on these indices, we can then determine which block from each input table is needed to carry out the computation.

Our scope ordering heuristic guarantees that, for any function $f(X_1, ..., X_k, Y)$, where $Y$ is the variable being eliminated, given any assignment of values to $X_1, ..., X_k$, all entries of the table corresponding to all different assignments to $Y$ reside within the same block. This implies that when a thread computes an entry in the table of a bucket's output function, it needs exactly one block from each of the input tables (see *Computation.2.d*).

**Proposition 5.** *Algorithm BEEM, given in Figure 6, is correct in terms of computing $P(e)$. It performs the same amount of work, $O(n \cdot k^{w^*})$, as regular BE, where $n$ is the number of variables, $k$ is variable domain size and $w^*$ is the induced width.* □

## 4 Experimental Evaluation

To evaluate the BEEM algorithm, we compared its performance with that of two other algorithms on computing the probability of evidence on probabilistic networks (Bayesian and Markov). In our comparison, we used a set of problems (called linkage/pedigree problems) derived from genetics. These problems were used

Algorithm **BEEM** for P(e)

**Input:** A problem description $P = <X, D, F>$; Evidence variables $var(e)$; A variable ordering $d = (X_1, ..., X_n)$ and its corresponding bucket-tree, $T = (X, E)$. The number of threads $m$ and main memory size $M$.

**Output:** P(e).

In the following, let

- $f_X$ be the function computed by bucket $B_X$, $(X, i)$ denote block $i$ of $f_X$ and $s_{(X,i)}$ ($e_{(X,i)}$) be the index of the first (last) element in block $(X, i)$,

- $nb_X$ denote the number of blocks in $f_X$ and $nc_X$ denote the number of blocks in $f_X$ that are computed,

- $rl_{(X,i)}$ be the list of threads using block $(X, i)$. Whenever a thread is removed from $rl_{(X,i)}$ and $rl_{(X,i)} = \emptyset$ as a result, block $(X, i)$ will be deleted. Also, whenever a thread is added to $rl_{(X,i)}$ and $rl_{(X,i)} = \emptyset$ before, block $(X, i)$ will be loaded from a disk file named "X-i".

- **Preprocessing:**

  1. Order the variables in the scope of each function as guided by the bucket-tree.

  2. Compute table block sizes $nb_X$ using Algorithm in Fig 5. This determines values $s_{(X,i)}$, $e_{(X,i)}$.

  3. **Initialize**: set all $nc_X = 0$, $rl_{(X,i)} = \emptyset$.

- **Computation**: in parallel, on $m$ threads, execute,

  1. Select a block to compute, from a bucket that has no non-computed children, resulting in a bottom-up computation order:

     Let $B_Y$ be a bucket in the bucket tree such that $nc_Y < nb_Y$ and for $\forall j \in C(Y)$, $nc_j = nb_j$, and $i$ be index of a block in $f_Y$ not yet computed. Set $nc_Y = nc_Y + 1$; mark $(X, i)$ as being computed.

  2. Enumerate all entries of block $(X, i)$:
     **for** $k = s_{(Y,i)}$ **to** $e_{(Y,i)}$,

     (a) Let $A$ be the assignment of values to arguments of $f_Y$ corresponding to $k$.
     (b) $\forall j \in C(Y)$, let $i_{j,k}$ be the index of the block, in $f_j$, corresponding to assignment $A \cup \{Y = 0\}$,
     (c) if $\exists j, i_{j,k} \neq i_{j,k-1}$ (meaning a current block held in memory for this child is changing),
        i. remove this thread from $rl_{(j,i_{j,k-1})}$,
        ii. add this thread to $rl_{(j,i_{j,k})}$.
     (d) compute entry $k$ of block $(Y, i)$ using the product-sum rule and input blocks $(j, i_{j,k})$.

  3. save block $(Y, i)$ to a disk file named "Y-i".

- **Return:** P(e) computed at root of the bucket-tree.

Figure 6: The BEEM algorithm for P(e)

in the solver competition held at the UAI-2008 conference[4]. Many of these problems were not solved in that competition and, in addition, we also consider a class of problems - type 4 linkage - with high induced width and large numbers of variables.

The two algorithms we used for comparison were: 1) VEC (Variable Elimination and Conditioning) [3]; and ACE[1]. Both VEC and ACE participated in the UAI-2008 solver competition. In their class (exact solvers for probabilistic networks, solving the P(e) problem) VEC/ACE were the two best solvers at the UAI-2008 competition, so a comparison with them seems warranted. A brief description of these algorithms follows in the next section.

Our experiments were carried out on a PC with an Intel quad-core processor, using 4 X 2TB hard disks in RAID-5 configuration (a total of 6TB of disk space). BEEM was configured so that $m = 4$ worker threads. As a space-saving measure, we deleted all input functions to a bucket after its output function was computed. All algorithms were given 1GB of RAM (main) memory and both VEC and BEEM were given the same variable orderings.

### 4.1 VEC

VEC is an algorithm that uses conditioning to produce sub-problems with induced widths small enough to be solved by an elimination algorithm[5]. A basic outline of VEC is as follows:

- As a pre-processing step, VEC reduces variable domains by converting all 0-probabilities to a SAT problem $F$ and checks for each assignment $X = a$ whether ($F$ and $X = a$) is consistent. Inconsistent assignments are pruned from the domain of $X$.

- Repeatedly, remove conditioning variables from the problem until the remaining problem fits within 1GB of main memory.

- Enumerate all value combinations of the conditioning variables. For each assignment, solve the remaining problem using variable elimination. Combine conditioned subproblem solutions to yield a solution to the entire problem.

### 4.2 ACE

ACE is a software package for performing exact inference on Bayesian networks developed in the Au-tomated Reasoning Group at UCLA[6]. ACE operates by compiling a Bayesian network into an Arithmetic Circuit (AC) and then using this AC to execute queries. Compiling into an AC occurs by first encoding the Bayesian network into Conjunctive Normal Form (CNF) and then extracting the AC from the factored CNF [1]. Encoding a network in this way efficiently exploits determinism, allowing ACE to answer queries on large networks in the UAI-08 solver competition.

---
[4]For a report on the results of the competition, see http://graphmod.ics.uci.edu/uai08/Evaluation/Report.
[5]see http://graphmod.ics.uci.edu/group/Software for more detail on VEC

### 4.3 Results

Preliminary results from running the three algorithms on a single class of problems are shown in Tables 1 and 2. In these tables $N$ indicates the number of variables, $w^\star$ is an upper bound on the induced width (determined experimentally using several min-fill orderings) and $K$ is the maximum domain size. The run time is presented in hh:mm:ss format and '>24 h' indicates the algorithm failed to compute p(e) in 24 hours, while 'OOM' indicates the algorithm exceeded the allotted 1 GB of RAM.

Table 1 contains results on pedigree problems with a few hundred to a thousand variables. Table 2 contains results from a set of problems with several thousand variables. On both sets of problems, we observe a few interesting phenomena. First, if a problem has $w^\star$ small enough that the problem fits into memory, all three algorithms compute p(e) very rapidly. In such situations, VEC and ACE may actually outperform BEEM because of the overhead associated with multi-threading. However, only BEEM and VEC are capable of solving problems that do not fit into RAM. In such situations, we see that the cost associated with reading and writing to hard disk is far less than the cost of conditioning. Finally, BEEM successfully computed p(E) for problems 7, 9, 13, 31, 34, 41, 50 and 51 for which an exact solution is not known.

## 5 Conclusions

We proposed an extension of the Bucket Elimination algorithm that utilizes external disk space for storing intermediate function tables. Extending the BE algorithm in this manner and also parallelizing computation is a non-trivial matter. In this paper we identified and addressed a number of key issues, including the decomposition of functions into appropriately sized blocks and processing to minimize access to hard disk.

While the performance of our algorithm is not fully optimized, it has shown very promising results on a class of large probabilistic networks. The algorithm demonstrates improved scalability, allowing exact computation of p(e) on problems not before solved by a general-purpose algorithm. To better understand its performance, we plan to run BEEM on several additional classes of problems. In addition to further improving the table decomposition and computation schemes, we also plan to extend BEEM for belief updating on variables other than the root and to handle more general tree decompositions. As illustrated in this paper, such modifications will inevitably impact the way in which tables are decomposed and processed.

### Acknowledgements

This work was supported in part by the NSF under award number IIS-0713118 and by the NIH grant R01-HG004175-02.

---
[6]http://reasoning.cs.ucla.edu/

| Problem | $w^*$ | N | max K | Space | BEEM runtime | VEC runtime | ACE runtime |
| --- | --- | --- | --- | --- | --- | --- | --- |
| pedigree1 | 15 | 334 | 4 | 37MB | 0:00:04 | 0:00:03 | 0:00:02 |
| pedigree7 | 30 | 1068 | 4 | 1,607,510MB | 9:50:09 | >24h | OOM |
| pedigree9 | 25 | 1118 | 7 | 10,192MB | 0:02:14 | >24h | OOM |
| pedigree13 | 31 | 1077 | 3 | 637,157MB | 3:29:35 | >24h | OOM |
| pedigree18 | 19 | 1184 | 5 | 166MB | 0:00:06 | 0:00:19 | 0:00:06 |
| pedigree19 | 23 | 793 | 5 | 149,928MB | 0:47:53 | >24h | OOM |
| pedigree20 | 22 | 437 | 5 | 16,510MB | 0:03:42 | 0:19:18 | 0:06:55 |
| pedigree23 | 25 | 402 | 5 | 50,301MB | 0:17:30 | 1:12:13 | 0:00:02 |
| pedigree25 | 23 | 1289 | 5 | 2,257MB | 0:00:31 | 0:01:40 | OOM |
| pedigree30 | 21 | 1289 | 5 | 563MB | 0:00:12 | 0:01:37 | OOM |
| pedigree31 | 30 | 1183 | 5 | 1,816,694MB | 9:34:54 | >24h | OOM |
| pedigree33 | 25 | 799 | 4 | 18,862MB | 0:04:38 | >24h | OOM |
| pedigree34 | 28 | 1160 | 5 | 657,895MB | 4:43:57 | >24h | OOM |
| pedigree37 | 20 | 1033 | 5 | 233,002MB | 1:10:40 | 0:05:46 | 0:00:56 |
| pedigree38 | 16 | 724 | 5 | 138,650MB | 0:47:59 | 0:07:23 | 0:06:25 |
| pedigree39 | 20 | 1272 | 5 | 900MB | 0:00:15 | 0:00:54 | OOM |
| pedigree40 | 29 | 1030 | 7 | 43,107,736MB | OOD | >24h | OOM |
| pedigree41 | 29 | 1062 | 5 | 881,226MB | 5:00:53 | >24h | OOM |
| pedigree42 | 22 | 448 | 5 | 61,149MB | 0:16:04 | >24h | OOM |
| pedigree44 | 25 | 811 | 4 | 19,187MB | 0:04:42 | 14:20:20 | OOM |
| pedigree50 | 17 | 514 | 6 | 1,465,526MB | 8:55:34 | >24h | OOM |
| pedigree51 | 35 | 1152 | 5 | 11,950,172MB | 76:35:16 | >24h | OOM |

Table 1: Time performance of BEEM, VEC and ACE on pedigree linkage problems. 'OOM' denotes Out of Memory, 'OOD' denotes Out of Disk and '>24h' indicates that the algorithm was stopped after 24 hours had elapsed. The BEEM algorithm was able to successfully compute p(e) on problems 7, 9, 13, 31, 34, 41, 50, 51 which had not been solved before.

| Problem | $w^*$ | N | max K | Space | BEEM runtime | VEC runtime | ACE runtime |
| --- | --- | --- | --- | --- | --- | --- | --- |
| type 4 100-16 | 30 | 6969 | 5 | 773,884MB | 4:20:10 | >24h | OOM |
| type 4 100-19 | 29 | 7308 | 5 | 149,731MB | 0:38:50 | >24h | OOM |
| type 4 120-17 | 24 | 7766 | 5 | 13,062MB | 0:02:56 | >24h | OOM |
| type 4 130-21 | 29 | 8844 | 5 | 107,434MB | 0:26:12 | >24h | OOM |
| type 4 140-19 | 30 | 9274 | 5 | 847,057MB | 5:04:47 | >24h | OOM |
| type 4 140-20 | 30 | 9355 | 5 | 2,555,753MB | 6:07:04 | >24h | OOM |
| type 4 150-14 | 32 | 9449 | 5 | 1,051,421MB | 6:16:58 | >24h | OOM |
| type 4 150-15 | 30 | 8290 | 5 | 2,614,816MB | 15:49:02 | >24h | OOM |
| type 4 160-14 | 31 | 10644 | 5 | 891,992MB | 4:46:16 | >24h | OOM |
| type 4 160-23 | 33 | 13514 | 5 | 3,359,257MB | 20:22:31 | >24h | OOM |
| type 4 170-23 | 21 | 11451 | 5 | 567MB | 0:00:36 | 0:04:59 | OOM |
| type 4 190-20 | 29 | 15765 | 5 | 389,000MB | 1:58:22 | >24h | OOM |

Table 2: Results of running BEEM, VEC and ACE on type-4 linkage problems.